\documentclass[journal]{IEEEtran} 
\IEEEoverridecommandlockouts
\usepackage{cite}
\usepackage{amsmath,amssymb,amsfonts}
\usepackage{algcompatible}
\usepackage{algorithm}
\usepackage{graphicx}
\usepackage{textcomp}
\usepackage{xcolor}
\usepackage{multirow}
\usepackage{booktabs}
\def\BibTeX{{\rm B\kern-.05em{\sc i\kern-.025em b}\kern-.08em
    T\kern-.1667em\lower.7ex\hbox{E}\kern-.125emX}}
\begin{document}

\title{3D-CSAD: Untrained \underline{3D} \underline{A}nomaly \underline{D}etection for \underline{C}omplex Manufacturing \underline{S}urfaces\\
}

\author{
    \IEEEauthorblockN{Xuanming Cao$^{1}$,
    Chengyu Tao$^{2}$, 
    Juan Du$^{1,3,4}$}
    \\
    \IEEEauthorblockA{$^{1}$ Smart Manufacturing Thrust, The Hong Kong University of Science and Technology (Guangzhou), Guangzhou, China}\\
    \IEEEauthorblockA{$^{2}$ Interdisciplinary
    Programs Office, The Hong Kong University of Science and Technology, Hong Kong SAR, China}\\
    \IEEEauthorblockA{$^{3}$ Department of
    Mechanical and Aerospace Engineering, The Hong Kong University of Science and Technology, Hong Kong SAR, China}\\
    \IEEEauthorblockA{$^{4}$ Guangzhou HKUST Fok Ying Tung Research Institute, Guangzhou, China}\\
}

\maketitle

\begin{abstract}
    The surface quality inspection of manufacturing parts based on 3D point cloud data has attracted increasing attention in recent years. The reason is that the 3D point cloud can capture the entire surface of manufacturing parts, unlike the previous practices that focus on some key product characteristics. However, achieving accurate 3D anomaly detection is challenging, due to the complex surfaces of manufacturing parts and the difficulty of collecting sufficient anomaly samples. To address these challenges, we propose a novel untrained anomaly detection method based on 3D point cloud data for complex manufacturing parts, which can achieve accurate anomaly detection in a single sample without training data. In the proposed framework, we transform an input sample into two sets of profiles along different directions. Based on one set of the profiles, a novel segmentation module is devised to segment the complex surface into multiple basic and simple components. In each component, another set of profiles, which have the nature of similar shapes, can be modeled as a low-rank matrix. Thus, accurate 3D anomaly detection can be achieved by using Robust Principal Component Analysis (RPCA) on these low-rank matrices. Extensive numerical experiments on different types of parts show that our method achieves promising results compared with the benchmark methods.
\end{abstract}


\section{Introduction}
Surface anomalies that occur during additive manufacturing or machining processes can have detrimental effects, leading to product failure and financial loss for manufacturers. Therefore, surface quality inspection plays a critical role in the manufacturing industry, to ensure that products meet the required standards and specifications. However, the current practice of anomaly detection relies heavily on the expertise and capabilities of human inspectors, which can be time-consuming, subjective, and prone to errors \cite{wang2021surface}. Thus, there is an urgent demand to develop effective and automatic defect detection algorithms to replace human labor.

Recently, the inspection of the entire surface quality is getting increasing attention thanks to the advance of 3D scanning technology, unlike the previous practices which only focus on a few key product characteristics of the manufacturing parts \cite{tao2023anomaly}. Compared with 2D images, 3D point cloud data can provide more information like depth and orientation of anomalies \cite{jovanvcevic20173d}. Based on high-resolution and high-precision 3D point cloud data, a few methods have been proposed to achieve automatic and accurate surface anomaly detection of various manufacturing surfaces. Since anomaly samples tend to be rare and data labeling is costly in practical manufacturing \cite{cohen2022semi}, the mainstream for anomaly detection research nowadays focuses on unsupervised learning and untrained methods. Among these, untrained method can directly deal with anomalous samples without any training on annotated or anomaly-free dataset, which has great potential for application in modern manufacturing.

\begin{figure}[t]
    \centerline{\includegraphics[width = \linewidth]{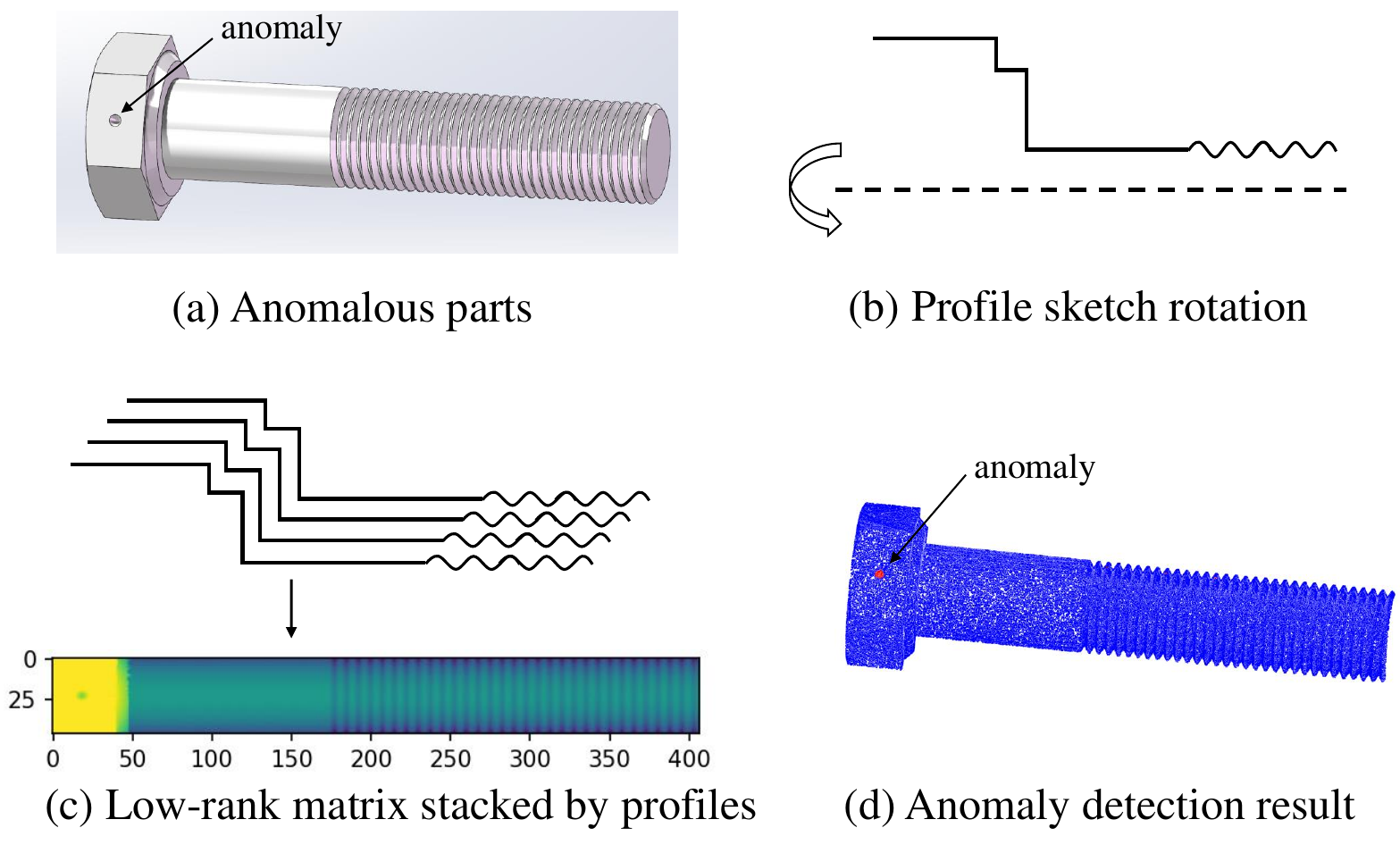}}
    \caption{Motivation of proposed 3D anomaly detection for complex manufacturing surfaces with a single sample. (a) is the input parts. (b) is the surface generation process from the perspective of design. (c) is the low-rank matrix stacked by multiple profiles, and (d) is anomaly detection results based on one single point cloud sample. }
    \label{motivation}
\end{figure}

Most existing untrained research assumes simple reference surfaces, such as planes \cite{tang2009characterization}, or smooth surfaces \cite{jovanvcevic20173d,tao2023anomaly,tao2023pointsgrade}. Note that the reference surface refers to the nominal surface of a manufacturing part. These methods can be effective even in the case of without training samples. However, they are inapplicable to more complex surfaces that exhibit non-smoothness in manufacturing scenarios. To deal with this case, training-based unsupervised data-driven methods are proposed \cite{horwitz2023back,bergmann2023anomaly,rudolph2023asymmetric}, which explicitly learn the representations of complex surfaces by machine learning or statistical tools. However, a training dataset with sufficient data is required.

Therefore, this paper aims to propose a new untrained anomaly detection method, dealing with the critical problem of achieving accurate 3D anomaly detection of complex surfaces without collecting a training dataset beforehand. However, this goal is challenging due to the following two reasons:
\begin{itemize}  
    \item The direct modeling of a complex surface represented by a 3D point cloud is difficult. 
    \item The training of many machine learning and statistical models fails when only a single sample exists.   
\end{itemize}

\begin{figure}[t]
    \centerline{\includegraphics[width = 0.9\linewidth]{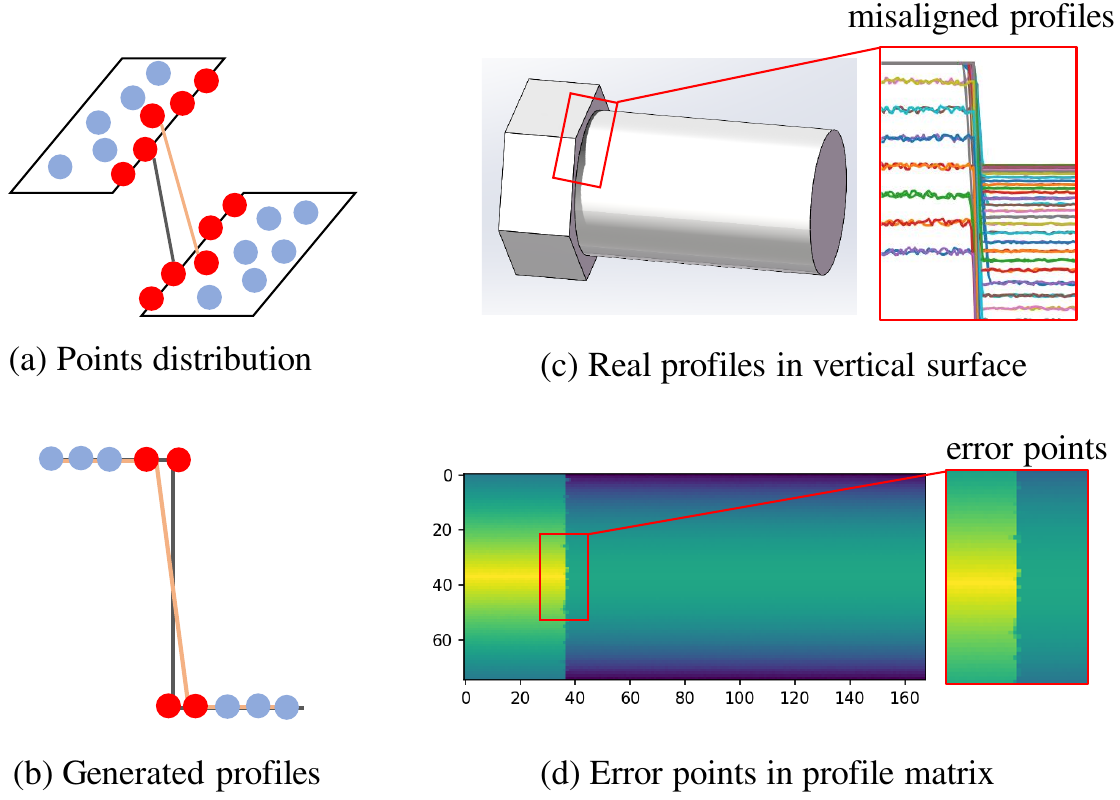}}
    \caption{Illustration of the causes of error points during profile generation. (a) The points distribution point cloud in the surface pointed to by the arrow, where blue points are the surface points and red points are edge points. (b) The generated deviated profile (in yellow) caused by unstructured edge points. (c) the real generated profiles. (d) The profile matrix with error points.}
    \label{outliers}
\end{figure}

To address these challenges, we propose a novel domain knowledge-infused approach, facilitating the modeling of a single complex manufacturing surface without the help of extra training data. Specifically, inspired by the design and manufacturing process of rotating bodies, i.e., the profiles of these parts have intrinsic shapes as shown in Fig. \ref{motivation} (b), we can transform a point cloud into a collection of profiles. These profiles share similar shapes, indicating that a few common bases are potentially sufficient to represent each profile. From a mathematical perspective, the matrix stacked by these profiles may be low-rank, as shown in Fig. \ref{motivation} (c). However, there are the following two issues in practice:
\begin{itemize}
    \item The profile generation is imperfect due to the randomly distributed points of unstructured point cloud data. As illustrated in Fig. \ref{outliers}, since the measured edge points are irregularly scattered, the generated profile (yellow in Fig. \ref{outliers} (b)) may deviate from the expected one (black in Fig. \ref{outliers} (b)). Then, the error points will occur in the profile matrix, as shown in Fig. \ref{outliers} (d), which correspond to the misalignment of profiles near edges in Fig. \ref{outliers} (c). 
    
    \item Even under the perfect profile generation, it is impossible that the low-rank assumption strictly holds on the entire surface, due to the complex nature of manufacturing parts.
\end{itemize}

    To tackle the above issues, we propose a new profile-based Component Segmentation and Cleaning module (CSC), considering that complex manufacturing parts usually comprise some basic components. By splitting the whole sample into multiple simpler components, the error points can be well eliminated. Simultaneously, for each component, the associated profiles are more likely to be low-rank, compared with the aforementioned entire surface. Finally, we adopt the Robust Principal Component Analysis (RPCA) \cite{candes2011robust} to cope with the processed components, enabling accurate sparse anomaly detection on the low-rank reference surface.

To sum up, this paper assumes that the focused manufacturing part is an axis-symmetric object, whose profiles have similar shapes. An example is the machining part as shown in Fig. \ref{motivation} (a). Moreover, the anomaly is assumed to be sparse, only occupying a small area on the surface.
With certain operations, the processed profiles lie in a low-rank space. Then, the RPCA algorithm can be used for accurate anomaly detection. The main contributions of this work are as follows:
\begin{itemize}
    \item We propose a novel domain knowledge-infused framework for 3D anomaly detection on complex manufacturing surfaces, which can deal with a single point cloud sample without requiring extra training data.
    \item We propose to transform a complex surface into multiple similar profiles inspired by the CAD design process, enabling complex reference surfaces to be modeled as low-rank representations.
    \item We devise a new CSC module for basic component segmentation and cleaning, and further adopt the RPCA on each component for accurate anomaly detection.
\end{itemize}

The remainder of this paper is organized as follows. Section \ref{related} introduces the related works about anomaly detection based on 3D point cloud data. Next, the proposed anomaly detection method for complex surfaces is described in detail in Section \ref{method}. After that, in Section \ref{exp}, extensive experiments on different manufacturing parts are conducted to validate our method. Finally, the conclusion of this paper is summarized in Section \ref{conclusion}.

\section{Related Works}
\label{related}
The rapid development of 3D scanning technology enables precise surface quality inspection, which attracts an increasing number of studies working on anomaly detection based on point cloud data. According to the requirements for data, this section reviews and classifies the existing works into two categories, i.e., Training-based and Untrained methods.

\subsection{Training-based methods}
Methods in this category require well-annotated data or sufficient normal sample data as training dataset, and can be further categorized into two subcategories.

\subsubsection{Supervised methods}
Well-annotated dataset is required in this type of methods for training machine learning classifiers to achieve anomaly detection. Madrigal et al. \cite{madrigal2017method} proposed a new local shape descriptor for 3D point cloud feature extraction, and utilized classical machining learning classifier, i.e., Support Vector Machines (SVM) to achieve anomaly detection. Chen et al. \cite{chen2021rapid} first clustered the input point cloud data into regions that potentially contain anomalies, based on which they then used a supervised classification algorithm to detect surface anomalies. Kaji et al. \cite{kaji2022deep} trained a deep neural network, i.e., RandLA-Net \cite{hu2020randla} to implement anomaly detection task.

\subsubsection{Unsupervised methods}
This type of approaches does not require data annotation, but requires a large number of normal samples to be included as training dataset. Von et al. \cite{von2016multiresolution} proposed to model the normal samples by using Principal Component Analysis (PCA) on an anomaly-free dataset, and then the anomalies can be identified by calculating the differences between the anomalous sample and the reconstructed one. Horwitz et al. \cite{horwitz2023back} stored all FPFHs from the normal samples in a memory bank. For the FPFHs of an anomalous sample, their distances from the collected memory bank were used for anomaly detection. Due to the large size of the memory bank,
this approach is time-consuming. To solve this problem, some deep learning-based algorithms have emerged \cite{bergmann2023anomaly, wang2023multimodal, rudolph2023asymmetric}. The typical one among them is the teacher-student network proposed by Bergmann et al. \cite{bergmann2023anomaly}, in which they compared the differences of representations from the teacher and student networks for anomaly detection, instead of constructing the memory bank. Notably, these data-driven methods do not impose any hypothesis on the shape of manufacturing parts. However, they heavily depend on the training dataset with sufficient samples to ensure effective learning processes. 

\subsection{Untrained methods}
The advantage of the approaches in this category is that no annotated data or in-control datasets are required for training, which means that they can achieve anomaly detection using a single sample without training.

\subsubsection{Nonmodel-based methods} Most methods belonging to this subcategory assume simple reference surfaces such as planes or smooth surfaces. Tang et al. \cite{tang2009characterization} fitted a global plane against all surface points as the ideal reference surface, and calculated the vertical deviation of each point from the fitted reference plane, in which the points with larger deviations than a given threshold were considered as anomaly points. Miao et al. \cite{miao2022pipeline} utilized a classical local shape descriptor, i.e., the Fast Point Feature Histogram (FPFH) \cite{rusu2009fast}, to model the reference surface of the turbine blade. Jovanvcevic et al. \cite{jovanvcevic20173d} designed a region-growing method to find the anomaly region in the point cloud surface, where the anomaly region was identified by comparing the difference of local geometric information, i.e. normal and curvature. Recently, Tao et al. \cite{tao2023anomaly} proposed a novel Bayesian network to model the unstructured point cloud data, in which the reference surface was modeled by a B-spline surface. 
The limitation of this type of approach is that they can only be applied to smooth or flat surfaces, incapable of dealing with more complex real-world manufacturing parts.

\subsubsection{Model-based methods} This subcategory assumes a given nominal model, e.g., the CAD model, as the reference surface. Then, anomaly detection is achieved by the comparison between the scanned point cloud and the nominal model through various registration techniques.
Decker et al. \cite{decker2020efficiently} devised a new engineering-informed registration technique for the quality assessment of 3D printed parts. Delacalle et al. \cite{delacalle2023hierarchical} presented a hierarchical registration method to reduce registration errors caused by dimensional defects or volumetric anomalies, achieving significant improvement in surface anomaly inspection. Yacob et al. \cite{yacob2019anomaly} used the deviation of skin model shape compared with the nominal surface to form histograms, based on which the anomaly sample can be detected. However, CAD models are not always available in practice, so these methods have limited applications.

Overall, there is currently very limited work investigating the 3D anomaly detection task for complex manufacturing parts without the requirement of the training dataset. The objective of this paper is to fill this gap by developing an untrained 3D anomaly detection method for complex manufacturing parts.

\begin{figure*}[t]
    \centerline{\includegraphics[width = \linewidth]{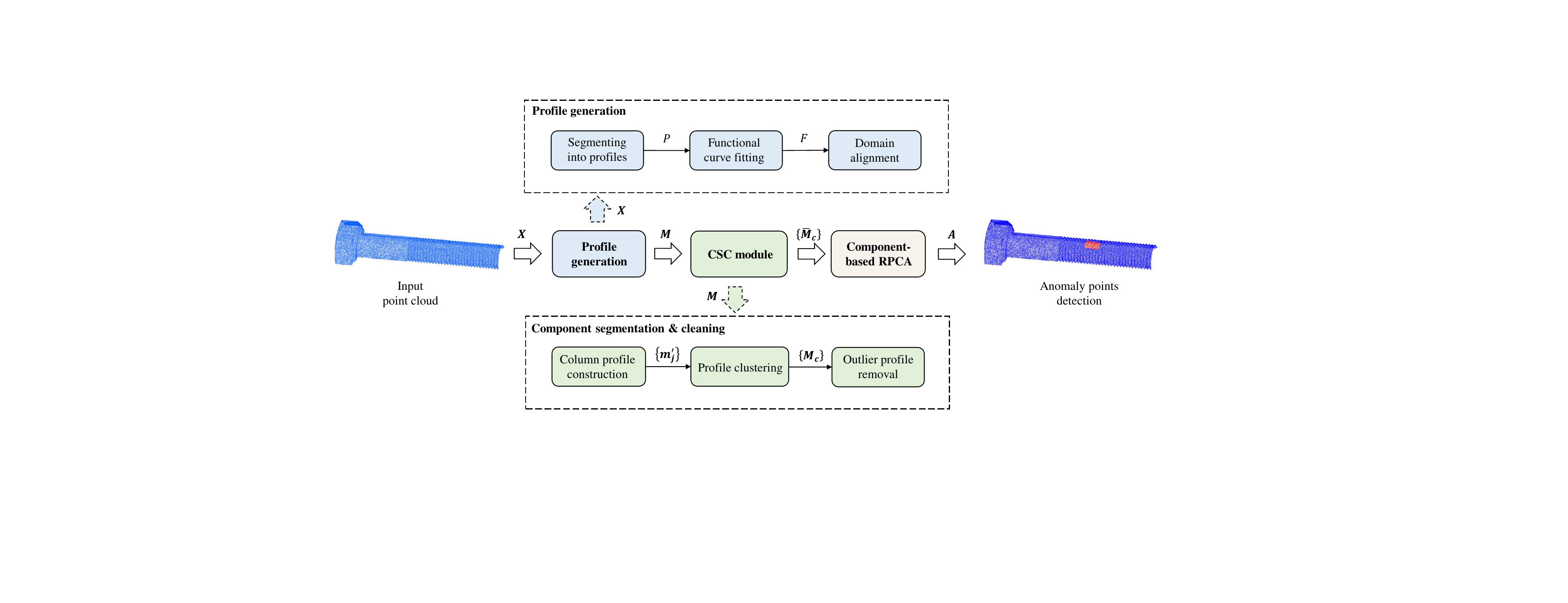}}
    \caption{The framework of the proposed method, which consists of three major steps, i.e., profile generation, component segmentation and cleaning module (CSC), and component-based RPCA (C-RPCA).}
    \label{framework}
\end{figure*}

\section{Methodology}
\label{method}
In this section, we will introduce the proposed methodology in detail. The goal is to achieve accurate 3D anomaly detection for complex manufacturing parts using only a single sample.

\subsection{Overview}
Our method consists of three major steps, which are Profile Generation, CSC Module, and Component-Based RPCA, respectively. Given a 3D point cloud $\mathbf{X} \in \mathbb{R}^{m \times 3}$ with $m$ points, Section \ref{sec:profile generation} first transforms it into a collection of profiles $\mathbf{M} \in \mathbb{R}^{n \times p}$ with similar shapes, where $n$ is the number of profiles and $p$ represents the number of points in each profile. To promote a low-rank representation and remove possible outliers of profiles, Section \ref{sec:csc} utilizes the CSC model to segment $\mathbf{M}$ into $c$ components $\left\{\bar{\mathbf{M}}_1, \ldots, \bar{\mathbf{M}}_c\right\}$, where each component $\bar{\mathbf{M}}_i \in \mathbb{R}^{n \times \bar{p}_i}$. Finally, the RPCA algorithm is operated on each $\bar{\mathbf{M}}_i$, stacking together into the final anomaly detection result of the input point cloud, as discussed in Section \ref{sec:component rpca}. The overview of our approach is illustrated in Fig. \ref{framework}.

\subsection{Profile Generation}
\label{sec:profile generation}
\begin{figure}[t]
    \centerline{\includegraphics[width = 0.9\linewidth]{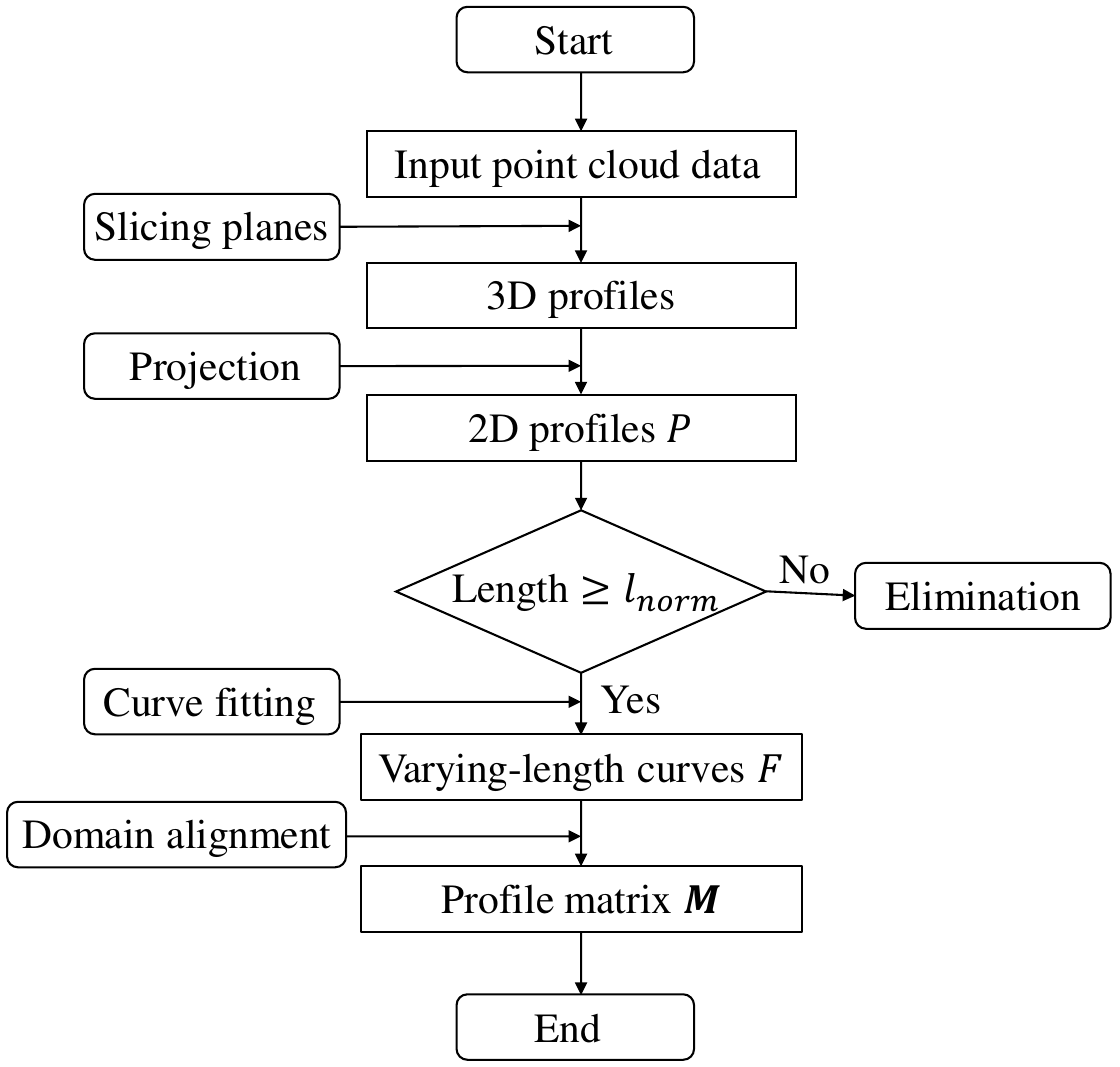}}
    \caption{The procedure of profile generation step.}
    \label{profile generate procedure}
\end{figure}

To model the complex reference surface as low-rank representations, the first step is to transform the input point cloud data into profiles, which consists of three steps, i.e., segmenting into profiles, functional curve fitting, and domain alignment. In this section, we will introduce them accordingly and the procedure is shown in Fig. \ref{profile generate procedure}. 

\begin{figure}[t]
    \centerline{\includegraphics[width = \linewidth]{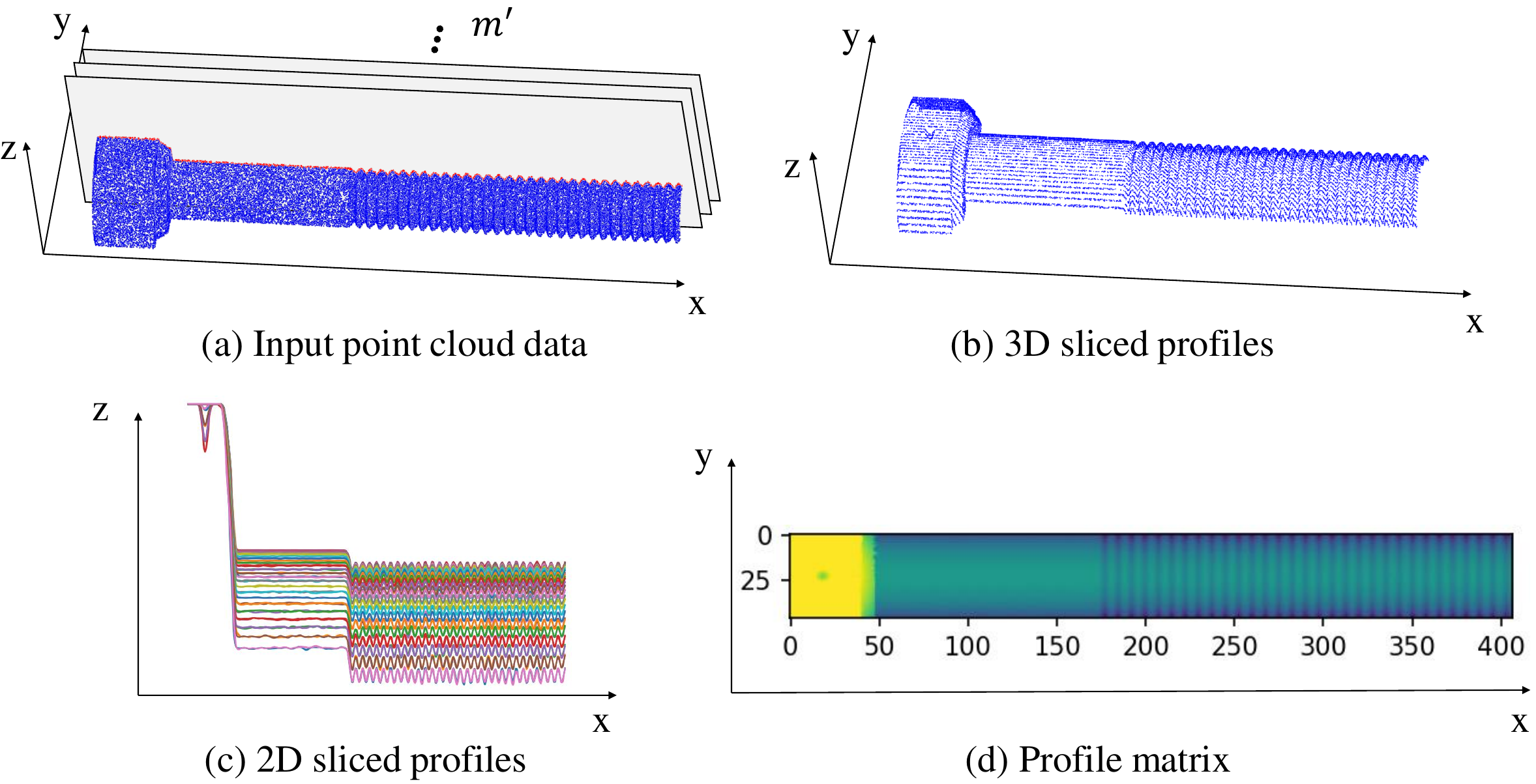}}
    \caption{The illustration of profile generation step. (a) the illustration of the slicing plane on input point cloud data. (b) the sliced 3D profiles. (c) 2D profiles fitted by B-spline curve. (d) the matrix stacked by the aligned 2D profiles.}
    \label{profile generate}
\end{figure}

\subsubsection{Segmenting into profiles} Concretely, as shown in Fig. \ref{profile generate} (a), $m^{\prime}$ parallel (to the $xz$ plane) slicing planes with the same interval $t$ are used to segment the original point cloud $\mathbf{X}$ into different 3D profiles. Each 3D profile is projected onto a corresponding slicing plane. The profiles with lengths smaller than $l_{\mathrm{norm}}$ are invaluable for anomaly detection and hence eliminated. Therefore, we generate a set of 2D profiles $P=\left\{\mathbf{p}_1, \ldots, \mathbf{p}_n\right\}$, $n<m^{\prime}$.

\subsubsection{Functional curve fitting} The original point cloud is unstructured, leading to the unordered and unequal-length profiles in $P$. To obtain a more regular data structure, we fit each profile into a cubic B-spline curve following the Functional Data Analysis (FDA) \cite{ramos2022scikit}, acquiring a set of functional curves $F=\left\{\mathbf{f}_1, \ldots, \mathbf{f}_{{n}}\right\}$.

\subsubsection{Domain alignment} To solve the unequal-length problem, a candidate is the extrapolation technique, which may be unreliable for complex surfaces. Instead, we simply truncate all functional curves into their common domain. This strategy is reasonable since the lengths of retained functional curves are close. In detail, denoting $d_{0,i}$ and $d_{1,i}$ as the lower bound and upper bound of the domain of the $i$th functional curve $\mathbf{f}_i$, the new domain for all functional curves is $[d_0, d_1]$, where $d_0 = \max \{d_{0,i}\}$ and $d_1 = \min \{d_{1,i}\}$. With the same domain, we can resample all functional curves into ordered discrete vectors $\left\{\mathbf{m}_{1}, \ldots, \mathbf{m}_{n} \in \mathbb{R}^{p}\right\}$, which are stacked to form the profile matrix $\mathbf{M}$, as illustrated in Fig. \ref{profile generate} (c) and (d).

\subsection{Component Segmentation \& Cleaning}
\label{sec:csc}

The obtained profiles after profile generation in Section \ref{sec:profile generation} have similar shapes. However, it is still problematic to directly apply RPCA on the profile matrix $\mathbf{M}$. As shown in Fig. \ref{outliers}, the unstructured nature of 3D point clouds introduces error points, which deviate more significantly than anomalies and are prone to be falsely detected. Furthermore, $\mathbf{M}$ constructed from the entire surface may not strictly satisfy the low-rank assumption, which needs to be segmented into more basic components. To address them, we propose a novel Component Segmentation \& Cleaning module (CSC) from the perspective of functional curve clustering, consisting of the following three steps, i.e., column profile construction, profile clustering, and outlier profile removal.

\subsubsection{Column profile construction} 
\begin{figure}[t]
    \centerline{\includegraphics[width = \linewidth]{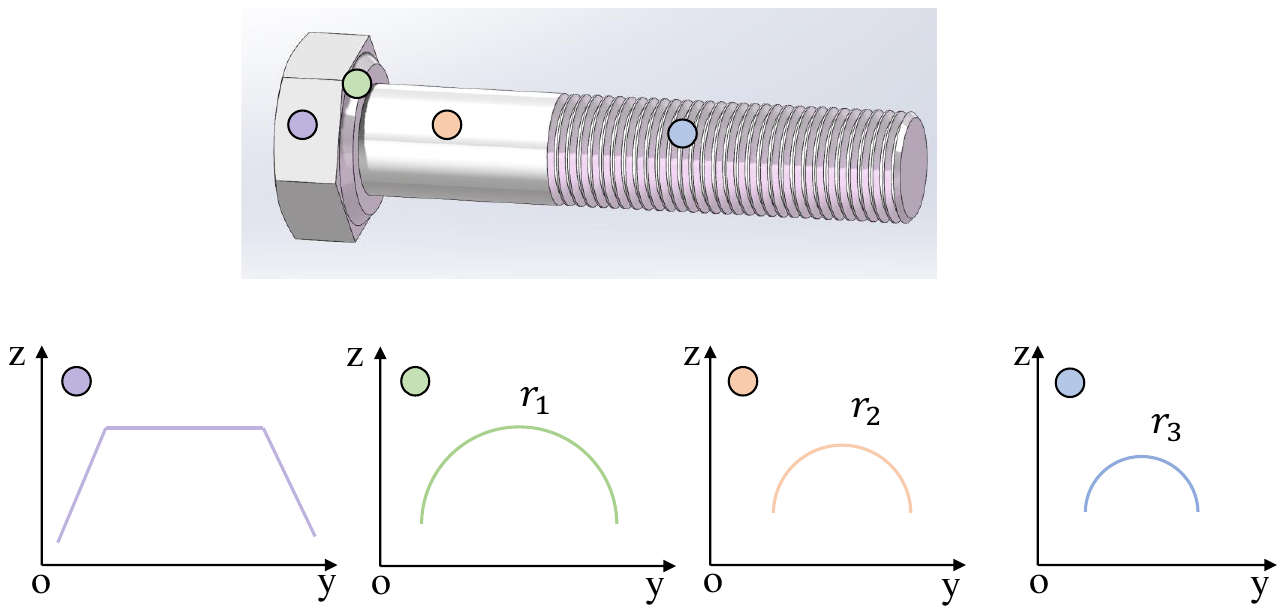}}
    \caption{Illustration of column profiles in different components. The bottom row shows the average shapes of column profiles in different components.}
    \label{column}
\end{figure}

Recall that the $i$th row of profile matrix $\mathbf{M} \in \mathbb{R}^{n \times p}$, i.e., $\mathbf{m}_i$, is a row profile or discrete vector. On the other hand, the $j$th column of $\mathbf{M}$, i.e., $\mathbf{m}^{\prime}_j$, can also be regarded as a column profile. For example, Fig. \ref{profile generate} and Fig. \ref{column} show that $\mathbf{m}_i$ and $\mathbf{m}^{\prime}_j$ are sliced along the $x$ and $y$ axes respectively. Different $\mathbf{m}_i$ shares similar shapes to each other, while the shape and dimension of $\mathbf{m}^{\prime}_j$ depend on its belonged component of the manufacturing part.
For example, as illustrated in Fig. \ref{column}, the bolt consists of four individual components. Then, $\mathbf{m}^{\prime}_j$ in the four components have different shapes or dimensions.

\subsubsection{Profile clustering} The above analysis indicates that the column profiles $\{\mathbf{m}^{\prime}_j\}$ can be clustered into $c$ groups, where $c$ is the number of basic components of a manufacturing part. To achieve this, we use the fuzzy c-means (FCM) for functional data clustering. Specifically, denoting the centroid of $k$th cluster as $\mathbf{c}_{k}$, the FCM is formulated as the following problem:
\begin{equation}
\begin{split}
    \min_{\{u_{ij}\},\{\mathbf{c}_{k}\}} \sum_{k=1}^c \sum_{j=1}^p u_{kj}^d\left\|\mathbf{m}^{\prime}_{j}-\mathbf{c}_k\right\|_2^2,\\
\text { s.t. } \sum_{i=1}^c u_{kj}=1, u_{kj} \geq 0 
\end{split}
\end{equation}
where $u_{kj}$ is associated with the probability of $\mathbf{m}^{\prime}_{j}$ belonging to the $k$th cluster, $d$ is the degree of fuzziness, and $\left\|\cdot\right\|_2$ is the $L_2$ norm. The details of the FCM procedure can be referred to \cite{ramos2022scikit}. The $c$ components of $\mathbf{M}$ are denoted as $\{\mathbf{M}_1,\ldots,\mathbf{M}_c\}$, where $\mathbf{M}_k\in \mathbb{R}^{n \times p_i}$ collects the column profiles in the $k$th cluster, and $\sum_{i=1}^c p_i=p$.

\subsubsection{Outlier profile removal} 
\begin{figure}[t]
    \centerline{\includegraphics[width = \linewidth]{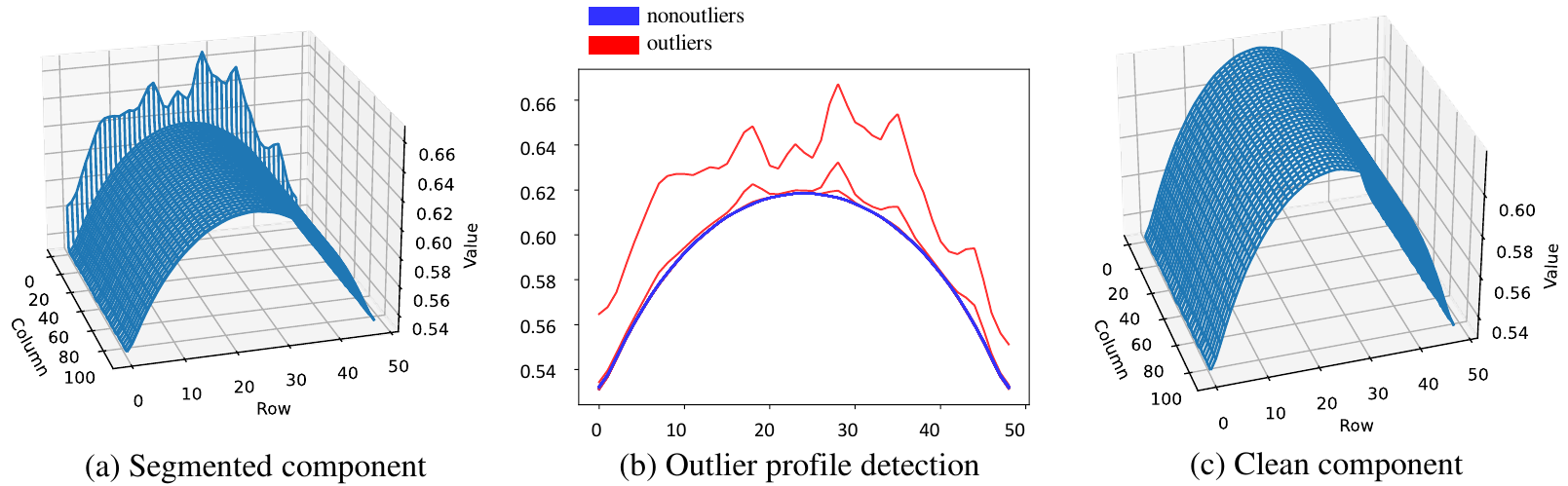}}
    \caption{The illustration of outlier profile removal. (a) is one of the segmented raw component in 3D view. (b) is the outlier profiles detection results. (c) is the clean component.}
    \label{outlier_remove}
\end{figure}

The error points in Fig. \ref{outliers} correspond to certain outlier profiles near component boundaries in each profile cluster. Therefore, for each cluster, we adopt the Functional Directional Outlyingness (FDO) technique \cite{dai2019directional} to remove outlier profiles, simultaneously considering the direction and magnitude of deviations between functional data and their central region. It is notable that we only remove the detected outlier profiles near the boundaries of a basic component, and keep the internal data that anomalies may fall into. This process is shown in Fig. \ref{outlier_remove}. Notably, the clean component $\bar{\mathbf{M}}_k \in \mathbb{R}^{n \times \bar{p}_k}$ removes the columns of $\mathbf{M}_k$ associated with the detected outlier profiles.

\subsection{Component-Based RPCA}
\label{sec:component rpca}

Instead of the profile matrix $\mathbf{M}$ of the entire surface, it is more suitable to apply the RPCA algorithm to each clean component $\bar{\mathbf{M}}_k$ processed by the CSC module, considering the simpler structure of a basic component. Following the basic idea of RPCA, the input matrix $\bar{\mathbf{M}}_k$ can be decomposed as a
low-rank reference surface $\mathbf{L}_k$ and the anomaly part $\mathbf{S}_k$ as follows:
\begin{equation}
\bar{\mathbf{M}}_k = \mathbf{L}_k  + \mathbf{S}_k .
\end{equation}
Then, the anomaly detection problem is to estimate the unknown anomaly part $\mathbf{S}_k$ by the following problem:
\begin{equation}
\min _{\mathbf{L}_k , \mathbf{S}_k , \bar{\mathbf{M}}_k  = \mathbf{L}_k  + \mathbf{S}_k }\|\mathbf{L}_k \|_*+\lambda\|\mathbf{S}_k \|_1,
\label{rpca}
\end{equation}
where $\|\mathbf{L}_k\|_*=\sum_r \sigma_r(\mathbf{L_k})$ is the nuclear norm used to ensure the low rank of $\mathbf{L}_k$ and $\sigma_r(\mathbf{L_k})$ takes the $r$-th singular value of $\mathbf{L}_k$. $\|\mathbf{S}_k\|_1=\sum_{ij} |\mathbf{S}_{k,ij}|$ is the $L_1$ norm to promote the sparsity of $\mathbf{S}_k$. Then the problem of Eq. \ref{rpca} becomes
\begin{equation}
    \min_{\mathbf{S}_k} \|\bar{\mathbf{M}}_k-\mathbf{S}_k\|_*+\lambda_k\|\mathbf{S}_k\|_1,
\end{equation}
where $\lambda_k$ is empirically defined by the sizes of data matrix $\bar{\mathbf{M}}_k$, i.e., $\lambda_k=(n + \bar{p}_i)/6$.
The anomalies $\{\mathbf{S_k}\}$ are concatenated into $\mathbf{S}\in \mathbb{R}^{n \times \sum_{k=1}^{c}{\bar{p}_k}}$ as follows:
\begin{equation}
\mathbf{S} = \mathbf{S_1} \oplus \mathbf{S_2} \oplus \ldots \oplus \mathbf{S_c},
\end{equation}
where $\oplus$ is the concatenation operation. Finally, $\mathbf{S}$ is mapped back to the anomaly score map $\mathbf{A}$ of the original point cloud $\mathbf{X}$ by finding the closest points between $\mathbf{X}$ and $\mathbf{S}$.

Finally, the complete procedure of our proposed methodology is summarized in Algorithm \ref{alg}.

\begin{algorithm}
\renewcommand{\algorithmicrequire}{\textbf{Input:}}
\renewcommand{\algorithmicensure}{\textbf{Output:}}
\caption{Anomaly detection for single complex manufacturing surface.}\label{alg}
\begin{algorithmic}[1]
\REQUIRE $\mathbf{X} \in \mathbb{R}^{m \times 3}$
\ENSURE $\mathbf{A}$
\State \underline{\textbf{// Profile generation}}
\State $P \gets \mathbf{X}$. Segmenting into profiles.
\State $F \gets P$. Functional curve fitting.
\State $\{d_{0,i}\}, \{d_{1,i}\} \gets F$. Domain ranges collection for all functional curves.
\State $d_{0}, d_{1}=\max \{d_{0,i}\}, \min \{d_{1,i}\}$. Domain alignment.
\State $\mathbf{M} \gets F$, over domain $[d_0, d_1]$. Points resampling on the aligned new domain.
\State \underline{\textbf{// CSC module}}
\State $\{\mathbf{m}^{\prime}_j\} \gets \mathbf{M}$. Profile matrix is split into column profiles.
\State $\{\mathbf{M}_1,\ldots,\mathbf{M}_c\} \gets \{\mathbf{m}^{\prime}_j\}$. Profile clustering using fuzzy-c means .
\State $\left\{\bar{\mathbf{M}}_1, \ldots, \bar{\mathbf{M}}_c\right\} \gets \{\mathbf{M}_1,\ldots,\mathbf{M}_c\}$. Eliminate the outlier profiles for each cluster.
\State \underline{\textbf{// Component-based RPCA}}
\FOR{$\bar{\mathbf{M}}_k$ in $\left\{\bar{\mathbf{M}}_1, \ldots, \bar{\mathbf{M}}_c\right\}$} 
\State $\mathbf{S_k} \gets RPCA(\bar{\mathbf{M}}_k)$. Component-based RPCA.
\ENDFOR
\State $\mathbf{S} = \mathbf{S_1} \oplus \mathbf{S_2} \oplus \ldots \oplus \mathbf{S_c}$. Anomalies concatenation.
\State $\mathbf{A} \gets \mathbf{S}$. Map anomaly scores back to the original input $\mathbf{X}$.
\end{algorithmic}
\end{algorithm}

\section{Experiments}
\label{exp}
In this section, we conduct extensive numerical experiments to validate our method and compare it with the state-of-the-art methods on different types of complex parts. Besides, the ablation study is performed to investigate the effectiveness and necessity of the key module in our method.

\subsection{Data Description} 
To evaluate the performance of the proposed approach, we utilize different types of manufacturing parts from simple to complex to generate reference point cloud surfaces, where \textit{hole} and \textit{scratch} anomalies are simulated as cones and planes edited in the 3D models.

\begin{figure}[t]
    \centerline{\includegraphics[width = \linewidth]{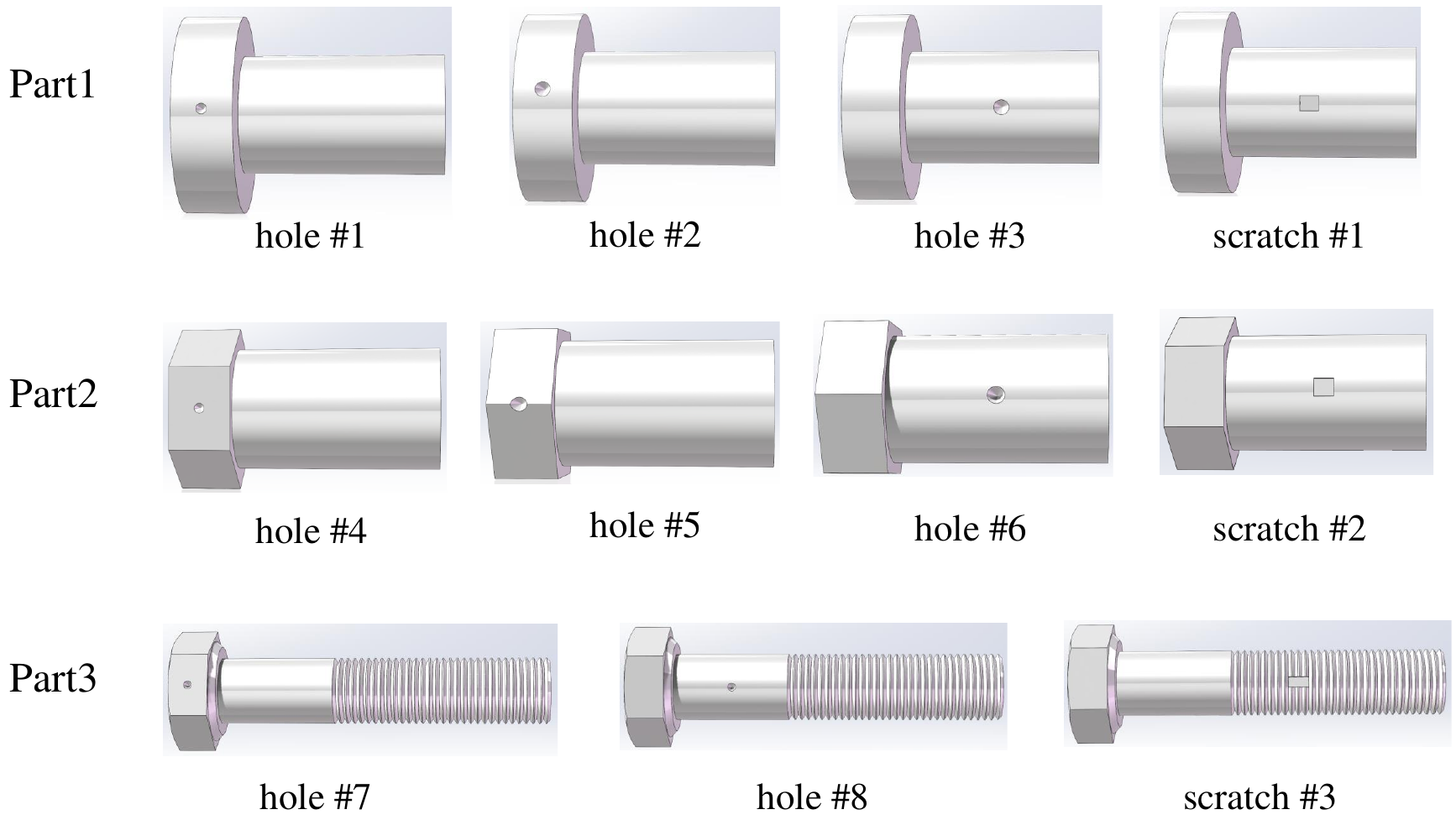}}
    \caption{The gallery of representative test samples. There are three types of parts with \textit{hole} and \textit{scratch} anomalies located in different positions.}
    \label{synthetic data}
\end{figure}

Specifically, we first generate the anomalies \textit{hole} and \textit{scratch} in the 3D models. For anomaly dimensions, the depth $d$ and radius $r$ of \textit{hole} anomalies are $d\in[0.9, 1.1], r\in[0.2, 0.5]$ respectively, and the length $l$, width $w$ and height $h$ of the \textit{scratch} anomalies are $l\in[0.6, 1.2], w\in[0.5, 1.1], h\in[0.1, 0.2]$. The bounding box sizes of Part1, Part2 and Part3 are (12.8, 10.0, 5.1), (12.8, 8.0, 3.4) and (28.5, 8.8, 3.6) respectively. Then, we obtain the 3D mesh from the anomalous CAD model and then sample the dense point cloud from the mesh data. For each point, the random noise $e_i$ which follows Gaussian distribution is added into it along the $x, y, z$ directions, and $\sigma_x, \sigma_y, \sigma_z$ are set as 0.001 for all samples. The gallery of representative test samples is shown in Fig. \ref{synthetic data}. Overall, for each anomaly category, there are 30 samples with different reference surfaces, different anomaly dimensions and locations, i.e., there are 60 test samples in total for validation, and the number of points for each point cloud sample is 100,000. The details can be seen in Table \ref{sample size}.

\begin{table}[]
\centering
\caption{Anomaly types and associated sample sizes.}
\label{sample size}
\setlength{\tabcolsep}{3mm}{
\begin{tabular}{@{}ccccc@{}}
\toprule
  Anomaly type      & Part1 & Part2 & Part3 & Total sample size \\ \midrule
hole    & 10    & 10    & 10    & 30            \\
scratch & 10    & 10    & 10    & 30            \\ \bottomrule
\end{tabular}}
\end{table}

\subsection{Evaluation Metrics}
Since pixel-wise anomaly detection can be considered as the binary classification for each point, the metrics used in classification are adopted in this work for comprehensive and objective evaluation. The metrics are accuracy (Acc), balanced accuracy (BA), Precision, Recall, and Dice coefficient (DICE), respectively. They are calculated by true positive (TP), false negative (FN), false positive (FP), and true negative (TN). In this paper, the TP indicates that the anomalous points are correctly detected as anomalies, FN represents that anomalous points are falsely identified as anomaly-free points, FP is that anomaly-free points are falsely detected as anomalies, and TN is that anomaly-free points are correctly identified as anomaly-free. 

\begin{equation}
\begin{aligned}
\mathrm{Precision} & =\frac{\mathrm{TP}}{\mathrm{TP}+\mathrm{FP}}, \quad \mathrm{Recall}=\frac{\mathrm{TP}}{\mathrm{TP}+\mathrm{FN}} \\
\mathrm{Acc} & =\frac{\mathrm{TP}+\mathrm{TN}}{\mathrm{TP}+\mathrm{FN}+\mathrm{FP}+\mathrm{TN}} \\ 
\mathrm{BA} & =\frac{\mathrm{TP}}{2(\mathrm{TP}+\mathrm{FN})}+\frac{\mathrm{TN}}{2(\mathrm{TN}+\mathrm{FP})} \\ 
\mathrm{DICE}&=\frac{2 \times \mathrm{Precision} \times \mathrm{Recall}}{\mathrm{PR}+\mathrm{Recall}}\\
\end{aligned}
\end{equation}
where Precision can measure the degree of normal points being falsely detected as anomalies, Recall measures whether all the anomaly points are identified, and DICE is the comprehensive metric that combines Precision and Recall. Obviously, the higher the value of these metrics, the better the performance of the algorithm.

\subsection{Comparison Results}

\begin{figure*}[t]
    \centerline{\includegraphics[width = \linewidth]{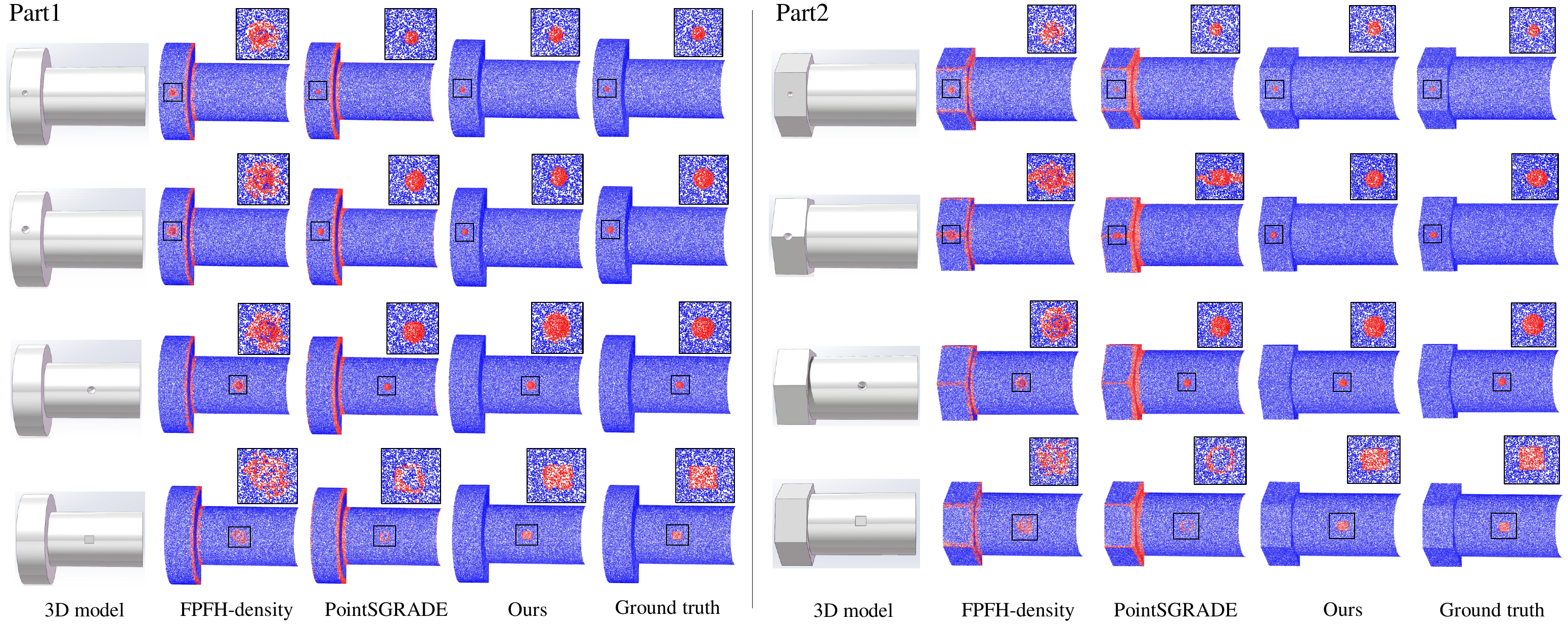}}
    \caption{Results of anomaly detection on representative samples of Part1 and Part2. The top three rows represent different locations and sizes of \textit{hole} anomaly, and the bottom row is results of \textit{scratch} anomaly (blue: normal point, red: detected anomaly point).}
    \label{part12}
\end{figure*}

As mentioned before, there are limited works about 3D anomaly detection for complex surfaces, among which Mohammadi et al. \cite{mohammadi2019non} propose to model the local shape features into density distribution as follows: 
\begin{equation}
f_h(x)=\frac{1}{n h} \sum_{i=1}^n K\left(\frac{x-x_i}{h}\right)
\end{equation}
where $K$ is the kernel smoothing function, $h$ is the bandwidth of the smoothing function. Potential anomaly points are determined based on the frequency of occurrence of each point.
Since they do not have the smooth assumption, we compare our method with this density-based method using the local shape features (FPFH), and we denote it as FPFH-density in the following parts. Additionally, a state-of-the-art method called PointSGRADE \cite{tao2023pointsgrade} that can handle free-form surfaces is also compared with our method. 

The parameter settings of each method are as follows:

\textbf{FPFH-density} \cite{mohammadi2019non}. The bandwidth of the smoothing function $h$ is determined by grid search, and the Gaussian kernel is used as the smoothing function. In addition, we set $k=30$ for the local shape descriptor FPFH, and downsample the number of points of the input point cloud data to 20,000, to balance the computation time and accuracy.

\textbf{PointSGRADE} \cite{tao2023pointsgrade}. The input sample is downsampled and scaled to the same level as \cite{tao2023pointsgrade}, and the parameters are set as recommended in the original paper, that is $k=800$, $k^{'}=1500$, $\lambda=0.007$, $\sigma=1.0$, and $\epsilon=0.001$.

\textbf{Ours}. The values and meaning of the parameters in our method are :
(1) $m^{'}$: the number of slicing planes, $m^{'}=100$.
(2) $l_{norm}$: the threshold of deleting the short profiles, $l_{norm}=0.8$. Note that we first normalize the input point cloud data, so that the maximum length of its bounding box is equal to 1.
(3) $p$: the number of resampled points of aligned profiles, $p=ratio \times 50$, where $ratio$ is the aspect ratio of the point cloud bounding box.
(4) $c$: the number of clusters for fuzzy c-means, $c=2$ for Part1 and Part2, $c=4$ for Part3.

\begin{figure}[t]
    \centerline{\includegraphics[width = \linewidth]{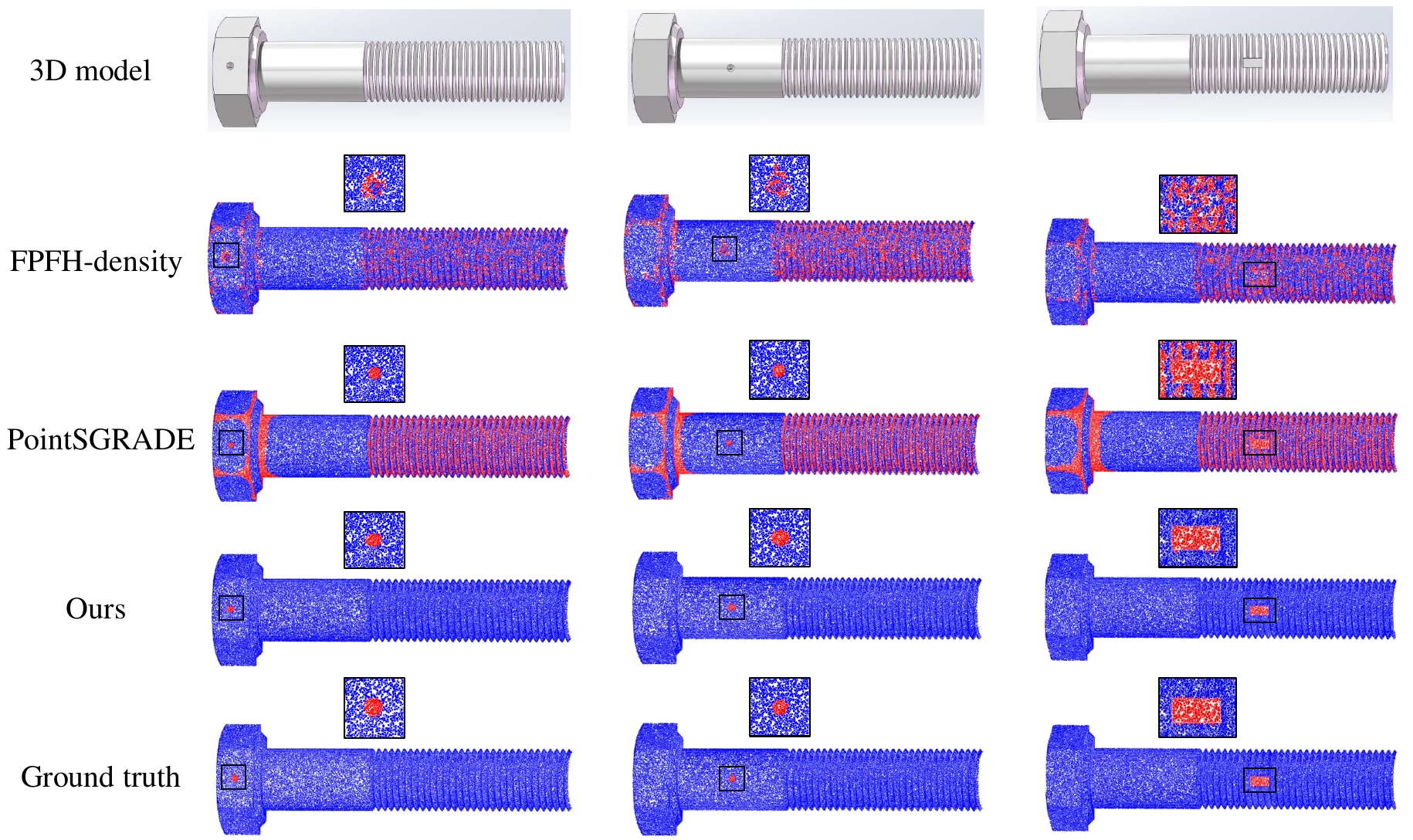}}
    \caption{Results of anomaly detection on representative samples of Part3, i.e., standard bolt, in which different columns show the detection results of different locations or types of anomalies.}
    \label{part3}
\end{figure}

The qualitative results of representative samples are shown in Fig. \ref{part12} and \ref{part3}, in which Fig. \ref{part12} gives the results of different approaches on Part1 and Part2 and Fig. \ref{part3} shows the comparison results of Part3, i.e., a standard bolt. We can see that the FPFH-density and PointSGRADE falsely detect the points in sharp edges and the thread region as anomaly points, which means that they cannot effectively handle the complex surface. In addition, PointSGRADE has better results than FPFH-density in terms of the \textit{hole} anomaly, because PointSGRADE is a reconstruction-based method that enables a fine anomaly boundary, while the anomaly boundary of FPFH-density can be blurred by the K-nearest neighbors. However, PointSGRADE cannot deal with the flat anomaly \textit{scratch}, since it assumes that the reference surface is the local plane. By contrast, our approach is not affected by the edge points or the complex thread region and correctly identifies the anomaly points whether they are \textit{hole} or \textit{scratch} anomalies compared with the ground truth. The reason is that the complex reference surface is modeled in our method by using RPCA, and only sparse anomalies can be detected.

\begin{table*}[t]
\caption{The quantitative results of anomaly detection}
\label{results tab}
\centering
\setlength{\tabcolsep}{5mm}{
\begin{tabular}{@{}cccccccc@{}}
\toprule
                       & Anomaly                  & Methods      & Acc             & BA              & Precision       & Recall          & DICE            \\ \midrule
\multirow{6}{*}{Part1} & \multirow{3}{*}{Hole}    & FPFH-density & 0.9288          & 0.8160          & 0.0560          & 0.7017          & 0.1028          \\
                       &                          & PointSGRADE  & 0.9000          & 0.9286          & 0.0545          & \textbf{0.9576} & 0.1026          \\
                       &                          & Ours         & \textbf{0.9992} & \textbf{0.9662} & \textbf{0.9475} & 0.9328          & \textbf{0.9390} \\ \cmidrule(l){3-8} 
                       & \multirow{3}{*}{Scratch} & FPFH-density & 0.9254          & 0.8077          & 0.0522          & 0.6886          & 0.0962          \\
                       &                          & PointSGRADE  & 0.8981          & 0.8514          & 0.0435          & 0.8041          & 0.0820          \\
                       &                          & Ours         & \textbf{0.9989} & \textbf{0.9521} & \textbf{0.9147} & \textbf{0.9048} & \textbf{0.9088} \\ \cmidrule(l){2-8} 
\multirow{6}{*}{Part2} & \multirow{3}{*}{Hole}    & FPFH-density & 0.9166          & 0.8234          & 0.0662          & 0.7284          & 0.1206          \\
                       &                          & PointSGRADE  & 0.8845          & 0.9212          & 0.0631          & \textbf{0.9586} & 0.1178          \\
                       &                          & Ours         & \textbf{0.9993} & \textbf{0.9755} & \textbf{0.9737} & 0.9513          & \textbf{0.9621} \\ \cmidrule(l){3-8} 
                       & \multirow{3}{*}{Scratch} & FPFH-density & 0.9096          & 0.8149          & 0.0523          & 0.7186          & 0.0963          \\
                       &                          & PointSGRADE  & 0.8817          & 0.8420          & 0.0444          & 0.8015          & 0.0834          \\
                       &                          & Ours         & \textbf{0.9988} & \textbf{0.9570} & \textbf{0.9214} & \textbf{0.9146} & \textbf{0.9176} \\ \cmidrule(l){2-8} 
\multirow{6}{*}{Part3} & \multirow{3}{*}{Hole}    & FPFH-density & 0.7002          & 0.5863          & 0.0049          & 0.4717          & 0.0097          \\
                       &                          & PointSGRADE  & 0.6485          & 0.7089          & 0.0068          & 0.7697          & 0.0135          \\
                       &                          & Ours         & \textbf{0.9995} & \textbf{0.9674} & \textbf{0.9334} & \textbf{0.9351} & \textbf{0.9340} \\ \cmidrule(l){3-8} 
                       & \multirow{3}{*}{Scratch} & FPFH-density & 0.7394          & 0.5784          & 0.0081          & 0.4158          & 0.0159          \\
                       &                          & PointSGRADE  & 0.6556          & 0.6634          & 0.0097          & 0.6713          & 0.0191          \\
                       &                          & Ours         & \textbf{0.9985} & \textbf{0.9297} & \textbf{0.8662} & \textbf{0.8602} & \textbf{0.8626} \\ \bottomrule
\end{tabular}}
\end{table*}

\begin{figure*}[t]
    \centerline{\includegraphics[width =0.8 \linewidth]{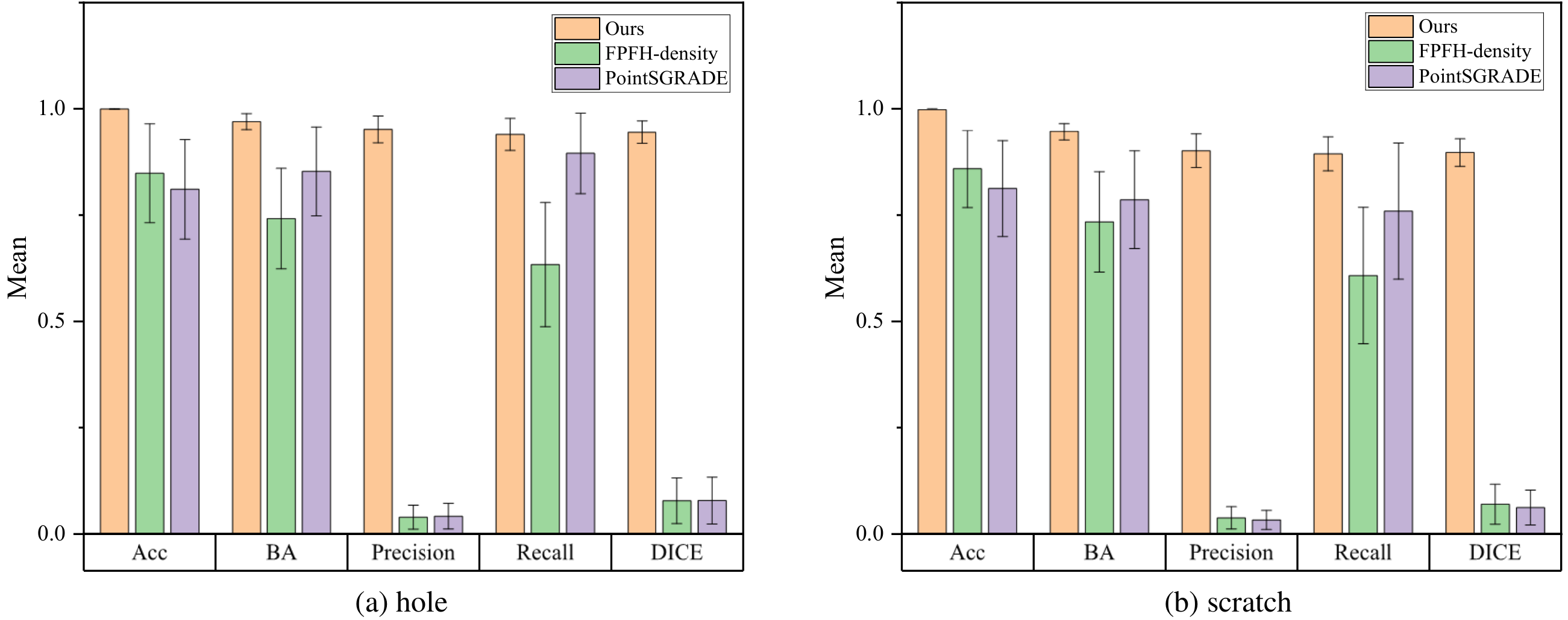}}
    \caption{Means and standard deviations of various evaluation metrics on different anomaly types of different methods. For each anomaly type, there are 30 samples for evaluation.}
    \label{error}
\end{figure*}

The quantitative results of different types of parts are illustrated in Table \ref{results tab}, whose values are the mean values of samples with different locations and sizes of anomalies. Fig. \ref{error} gives the means and standard deviations of various evaluation metrics on different anomaly types of different methods. Note that we tuned the thresholds to ensure the highest DICE values for each experiment. We can see that both FPFH-density and PointSGRADE have low Precision on all the test samples, due to the high false detection rate in the sharp edges and thread region. Also, PointSGRADE has larger values in terms of Recall than FPFH-density, since the anomaly boundaries are more accurately detected by PointSGRADE than those in FPFH-density. However, the comprehensive evaluation metric DICE of PointSGRADE is still disastrous, due to its low Precision values. By contrast, our method achieves the best results in terms of almost all the evaluation metrics and outperforms benchmark approaches by a large margin.

\subsection{Ablation Study}
In our framework, we propose the CSC module to address two problems, i.e., eliminating the error points and providing basic and simpler components for RPCA to better satisfy the low-rank assumption. In this module, there are two essential operations, including component segmentation (CS) and outlier profile removal (OPR), which enable accurate anomaly detection on complex point cloud samples. Therefore, we conduct ablation experiments to validate the necessity of these two operations in this section, and give the quantitative and qualitative results on Part2 with \textit{hole} anomaly. 

\begin{figure}[t]
    \centerline{\includegraphics[width = \linewidth]{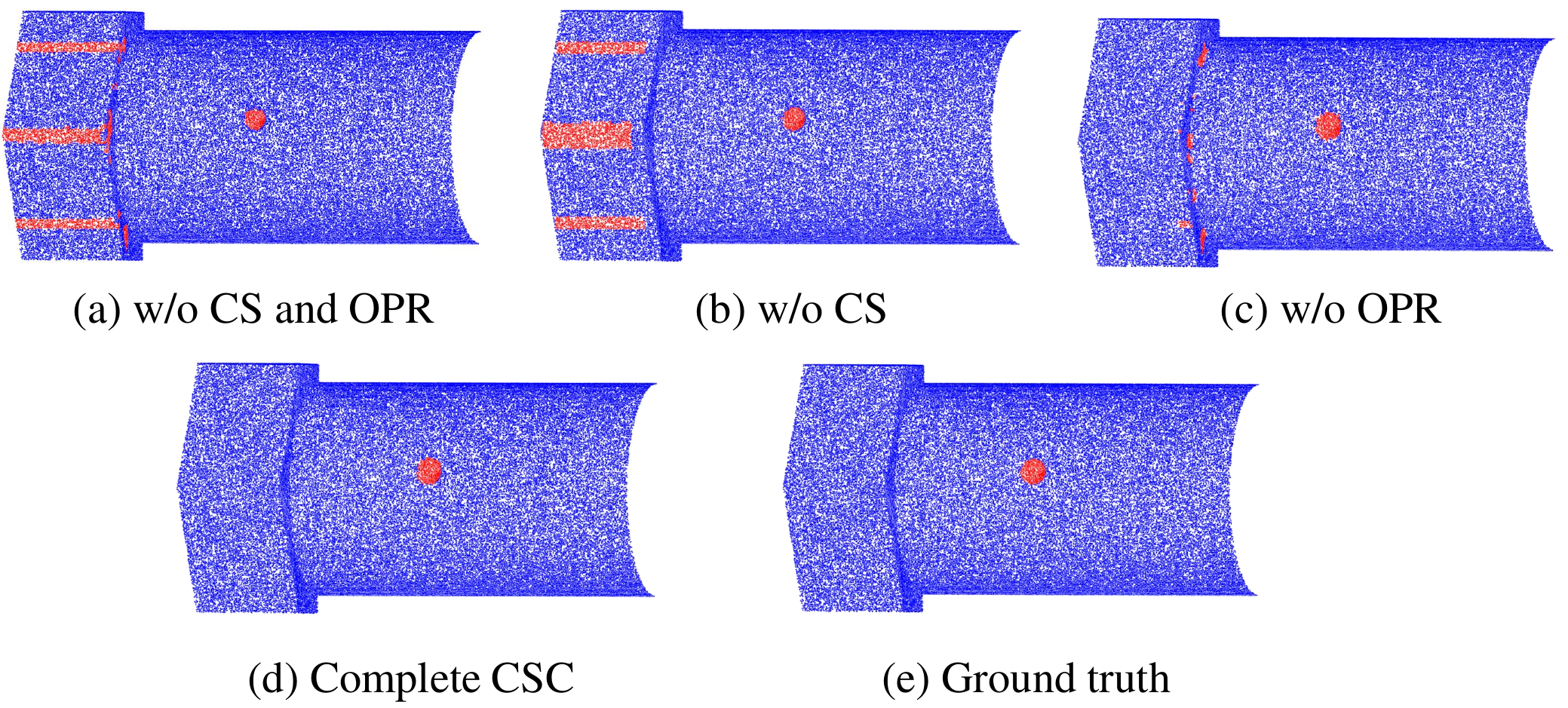}}
    \caption{Visualization of ablation study on a representative sample. (a) Anomaly detection result without component segmentation (CS) and outlier profile removal (OPR). (b) The result without component segmentation (CS). (c) The result without outlier profile removal (OPR). (d) The result with complete CSC module. (e) Ground truth.}
    \label{ablation}
\end{figure}

Fig. \ref{ablation} shows the results of removing the component
segmentation and outlier profile removal, respectively. If we remove the component segmentation step, there will be a large number of normal points being falsely detected as anomaly points, as illustrated in Fig. \ref{ablation} (b). This is because the matrix stacked by the whole profiles instead of segmented profiles does not satisfy the low-rank assumption, which causes the failure of RPCA. In addition, when we omit the outlier profile removal step, the error points are falsely detected as anomalies as demonstrated in Fig. \ref{ablation} (c). The reason is that error points tend to be sparse and have a larger deviation from the reference surface than anomaly points, which hinders accurate anomaly detection. Furthermore, Fig. \ref{ablation} (a) gives the result of removing both outlier profile removal and component segmentation steps, which provides much poorer performance.

The quantitative results of ablation experiments are shown in Table \ref{ablation tab}. As we can see, whether removing outlier profile removal or component segmentation step causes a large number of normal points to be falsely detected as anomalies, which leads to a significant reduction in the Precision values compared to the complete algorithm. Observing the DICE values, we can find that without the outlier profile removal step there is a less impact on the performance of the anomaly detection than without component segmentation step.

Therefore, we can conclude from ablation experiments that the proposed CSC module, including segmenting the sample into components and cleaning each component, is indispensable for achieving accurate anomaly detection in complex manufacturing surfaces.

In conclusion, our method achieves promising results on different types of complex samples and outperforms the competitors by a large margin. The superiority of the proposed algorithm is attributed to the following two factors: First, we mathematically model the complex reference surface of the test samples by using the low-rank representation, which significantly reduces the false detection rate. In addition, the CSC module provides basic and clean components for RPCA to achieve accurate anomaly detection.

\begin{table}[]
\caption{The quantitative results of ablation study on Part2 with \textit{hole} anomaly.}
\label{ablation tab}
\resizebox{\linewidth}{!}{
\begin{tabular}{@{}ccccccc@{}}
\toprule
\multicolumn{2}{c}{CSC module}                                                                                                         & \multirow{2}{*}{Acc} & \multirow{2}{*}{BA} & \multirow{2}{*}{Precision} & \multirow{2}{*}{Recall} & \multirow{2}{*}{DICE} \\
\begin{tabular}[c]{@{}c@{}}Outlier profile \\ removal\end{tabular} & \begin{tabular}[c]{@{}c@{}}Component \\ segmentation\end{tabular} &                       &                            &                         &                      &                     \\ \midrule
                                                                   &                                                                   & 0.9284                & 0.8271                     & 0.0964                  & 0.7240               & 0.1614              \\
\checkmark                                                                  &                                                                   & 0.9382                & 0.9170                     & 0.1187                  & 0.8953               & 0.2034              \\
                                                                   & \checkmark                                                                 & 0.9934                & 0.9115                     & 0.5885                  & 0.8281               & 0.6740              \\
\checkmark                                                                  & \checkmark                                                                 & \textbf{0.9993}                & \textbf{0.9755}                     & \textbf{0.9737}                  & \textbf{0.9513}               & \textbf{0.9621}              \\ \bottomrule
\end{tabular}}
\end{table}

\section{Conclusion}
\label{conclusion}
In this paper, we propose a novel anomaly detection framework for complex manufacturing surfaces based on 3D point cloud data. The proposed method transforms the complex sample into multiple similar profiles, to model the complex reference surface as a low-rank representation. Furthermore, to further satisfy the low-rank assumption and eliminate the possible error points caused by unstructured point cloud data, the CSC module is proposed, where the complex sample is first segmented into basic and simple components, after which the segmented components are cleaned by outlier profile removal step. Finally, RPCA is adopted on these processed clean components to decompose the reference surfaces and sparse anomalies, so that accurate anomaly detection can be achieved. Extensive comparison experiments on different types of samples show the superiority of our approach over the state-of-the-art methods, and the ablation study also demonstrates the effectiveness of the key module in our framework.



\vspace{12pt}

\bibliographystyle{plain}
\bibliography{ref}
\end{document}